\documentclass{article}

\PassOptionsToPackage{numbers, compress}{natbib}

\usepackage[final]{drlw_neurips_2019}

\usepackage[utf8]{inputenc} 
\usepackage[T1]{fontenc}    
\usepackage{hyperref}       
\usepackage{url}            
\usepackage{booktabs}       
\usepackage{amsfonts}       
\usepackage{nicefrac}       
\usepackage{microtype}      
\usepackage{amsmath}
\usepackage{subcaption}
\usepackage{graphicx}
\usepackage{float}
\title{Multiplayer AlphaZero}

\author{
  Nick Petosa \\
  School of Interactive Computing\\
  Georgia Institute of Technology\\
  \texttt{npetosa@gatech.edu} \\
   \And
   Tucker Balch \textsuperscript{1}\\
   School of Interactive Computing\\
Georgia Institute of Technology\\
   \texttt{tucker@cc.gatech.edu} \\
}

\begin{document}

\maketitle

\begin{abstract}
The AlphaZero algorithm has achieved superhuman performance in two-player, deterministic, zero-sum games where perfect information of the game state is available. This success has been demonstrated in Chess, Shogi, and Go where learning occurs solely through self-play. Many real-world applications (e.g., equity trading) require the consideration of a multiplayer environment. In this work, we suggest novel modifications of the AlphaZero algorithm to support multiplayer environments, and evaluate the approach in two simple 3-player games. Our experiments show that multiplayer AlphaZero learns successfully and consistently outperforms a competing approach: Monte Carlo tree search. These results suggest that our modified AlphaZero can learn effective strategies in multiplayer game scenarios. Our work supports the use of AlphaZero in multiplayer games and suggests future research for more complex environments.

\end{abstract}

\footnotetext[1]{On leave at J.P. Morgan AI Research.}
\footnotetext[2]{https://github.com/petosa/multiplayer-alphazero}

\section{Introduction}

DeepMind's AlphaZero algorithm is a general learning algorithm for training agents to master \textit{two-player, deterministic, zero-sum games of perfect information} \cite{deepmind}. Learning is done \textit{tabula rasa} - training examples are generated exclusively through self-play without the use of expert trajectories. Unlike its predecessor AlphaGo Zero, AlphaZero is designed to work across problem domains \cite{alphagozero}. DeepMind has demonstrated AlphaZero's generality by training state-of-the-art AlphaZero agents for Go, Shogi, and Chess. This result suggests that AlphaZero is applicable to other games and real-world challenges. In this paper, we explore AlphaZero's generality further by evaluating its performance on simple \textit{multiplayer} games.

Our approach is to extend the original two-player AlphaZero algorithm to support multiple players through novel modifications to its tree search and neural network architecture. Since the AlphaZero source code is not released, we implemented a single-threaded version of AlphaZero from scratch using Python 3 and PyTorch based on DeepMind's papers \cite{deepmind, alphagozero}. There are several notable reimplementations of DeepMind's AlphaGo Zero algorithm by the research community such as LeelaZero and ELF \cite{leela, elf}. However, these implementations are designed and optimized for reproducing DeepMind's results on Go, not for general experimentation with the algorithm.

Our contribution is threefold. First, we produce an independent reimplementation\textsuperscript{2} of DeepMind's AlphaZero algorithm. Second, we extend the original algorithm to support multiplayer games. And third, we present the empirical performance of this extended algorithm on two multiplayer games using some novel evaluation metrics. We conclude that the AlphaZero approach can succeed in multiplayer problems.

This paper will first introduce the original AlphaZero algorithm, then discuss our novel multiplayer extensions, and lastly discuss our experiments and results.

\section{Background}

The original, two-player AlphaZero can be understood as an algorithm that learns a board-quality heuristic to guide search over the game tree. This can be interpreted as acquiring an ``instinct'' for which board states and moves are likely to end in victory or defeat, and then leveraging that knowledge while computing the next move to make. The resulting \textit{informed search} can pick a high quality move in a fraction of the time and steps as an \textit{uninformed search}.

This ``instinct heuristic'' is the output of a deep convolutional neural network, which ingests the current board state as input and outputs two values \cite{alexnet}. The first output is the \textit{value head} ($v$), the scalar utility of this board from the perspective of the current player. The second output is the \textit{policy head} ($\vec{p}^{\,}$), a probability distribution over legal actions from the current board state, where higher probability actions should lead the current player to victory. Both $v$ and $\vec{p}^{\,}$ inform a Monte Carlo tree search (MCTS) to guide search over the game tree.

\subsection{Informed MCTS}

MCTS is a search algorithm that traverses the game tree in an exploration/exploitation fashion \cite{mcts}. At each state, it prioritizes making moves with high estimated utility, or that have not been well explored. The upper confidence bound for trees (UCT) heuristic is often used to balance exploration and exploitation during search \cite{uct}. Each iteration of MCTS from a board state is called a ``rollout.'' AlphaZero uses most of the standard MCTS algorithm, but with a few key changes.
\begin{enumerate}

    \item Replaces UCT with the following $(state, action)$-pair heuristic in MCTS to decide which move to search next.
    $$
    Q(s,a) + c_{puct}\frac{P(s,a)}{1+N(s,a)}
    $$
    Where $Q$ is the average reward experienced for this move, $N$ is the number of times this move has been taken, $P$ is the policy head value, and $c_{puct}$ is an exploration constant. The design of this heuristic trades off exploration of under-visited moves with exploitation of the value and policy heads from the network.
    \item Random rollouts are removed. Instead of rolling out to a terminal state, the value head $v$ is treated as the approximate value of a rollout. Because of this, during the backpropagation step of MCTS, $v$ gets incorporated into $Q$.

\end{enumerate}

The result is an ``informed MCTS'' which incorporates the outputs of the neural network to guide search.

\subsection{Policy iteration}

To train, AlphaZero operates in a cycle of policy evaluation and policy improvement. AlphaZero requires full access to a simulator of the environment.

\paragraph*{Policy evaluation}
Training proceeds as follows. Our game starts at initial board state ($s_1$). From here, several rollouts of MCTS are run to discover a probability distribution ($\vec{\pi_1}^{\,}$) across valid actions. Uniform dirichlet noise is added to $\vec{\pi_1}^{\,}$ to encourage exploration (this is only done for the first move of the game). We then take our turn by sampling from $\vec{\pi_1}^{\,}$ to get to $s_2$, and repeat the process until a terminal state is encountered. This terminal state outcome $z$ will be 1 for win, -1 for loss, or 0 for tie from the perspective of the current player.

\paragraph*{Policy improvement}
 After a game ends, we generate training samples for each turn of the game $(s_i, \vec{\pi_i}^{\,}, z)$ and add them to an experience replay buffer. After several games, we sample batches from the buffer to update our network parameters by minimizing the following loss function, which is just a sum of cross-entropy loss and mean squared error. 
 
 $$
 L=(z-v)^2 -  \vec{\pi}^{\,\top} \log{\vec{p}^{\,}}
 $$

Through this cycle, AlphaZero refines its heuristic after each iteration and snowballs into a strong player.

\section{Multiplayer extensions}
Several changes are made to MCTS and the neural network for AlphaZero to support multiplayer games.
\begin{enumerate}

\item MCTS now rotates over the full list of players during play instead of alternating between two players.

\item
Instead of completed games returning an outcome $z$, they now return a \textit{score vector} ($\vec{z}^{\,}$), indicating the scores of each player. For example, in a 3-player game of Tic-Tac-Toe, a tie might return $\begin{bmatrix} 0 & 0 & 0 \end{bmatrix}$ and a first player win might return $\begin{bmatrix} 1 & -1 & -1 \end{bmatrix}$. Note from the latter example that we are incidentally relaxing the zero-sum constraint on games. In fact, this opens the door for games that do not have binary win/lose outcomes, but this is not the focus of our work.

\item In two-player AlphaZero, the value of a state from the perspective of one player is the negation of the value for the other player. With a score vector, each player can have its own score. So when backpropagating value, MCTS uses the corresponding score in $\vec{v}^{\,}$ for each player instead of flipping the sign of a scalar $v$. 

\item
Instead of the value head of the neural network predicting a scalar value, it now predicts a \textit{value vector} ($\vec{v}^{\,}$), which contains the expected utility of a state for each player. The size of the vector equals the number of players in the game. The loss function is updated to account for this change.

 $$
 L=\frac{1}{n}\sum_{i=1}^n(z_i-v_i)^2 -  \vec{\pi}^{\,\top} \log{\vec{p}^{\,}}
 $$
 
 The neural network is now trained on $(s_i, \vec{\pi_i}^{\,}, \vec{z}^{\,})$ tuples since value is a vector. An illustration of this change from the standard two-player case to the novel multiplayer case is shown in figure \ref{fig:crownjewel} for a 3-player variant of Tic-Tac-Toe.
 
\begin{figure}[h]
\centering
\includegraphics[width=120mm]{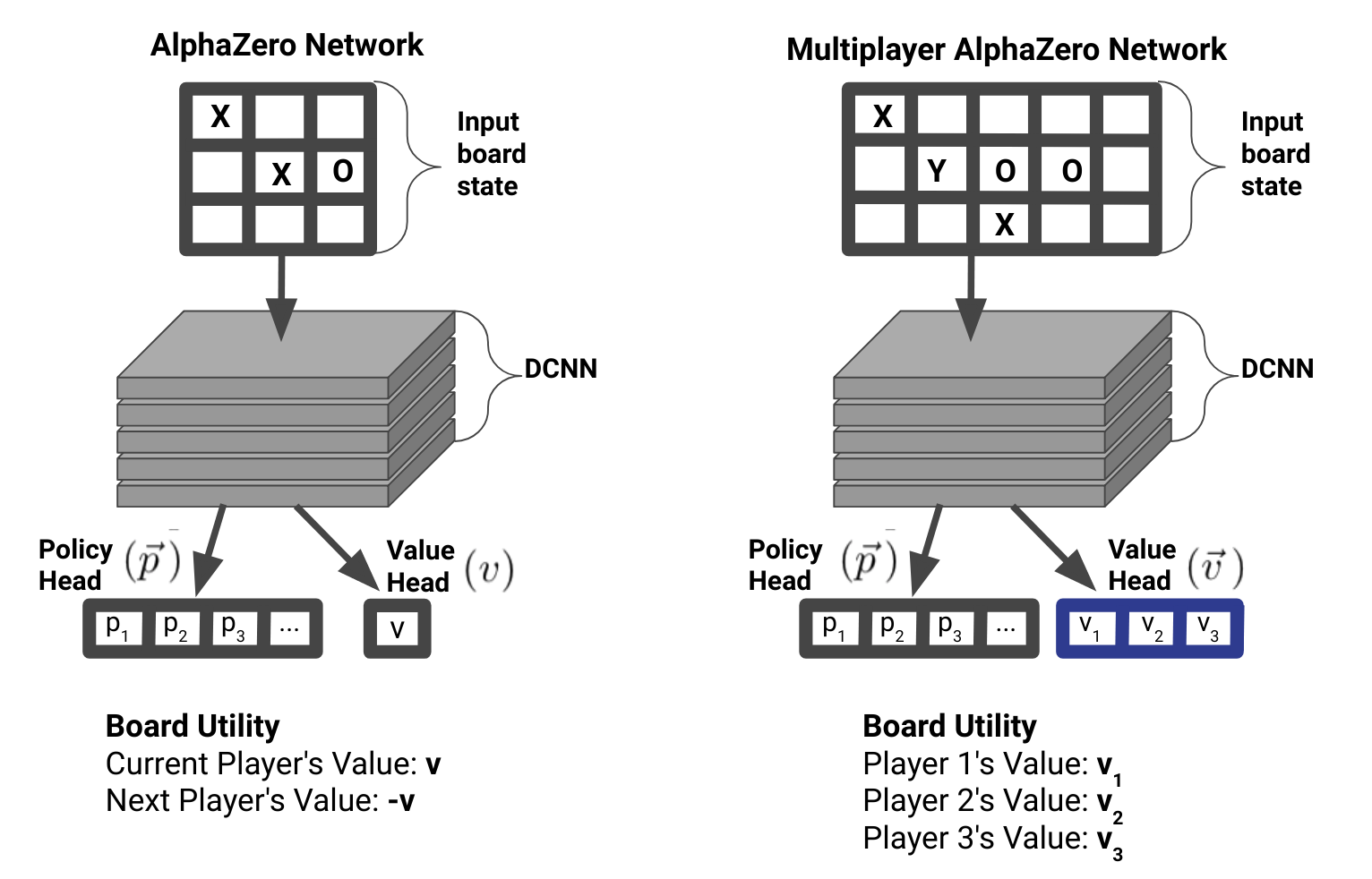}
\caption{The change in neural network structure with novel multiplayer approach.}
\label{fig:crownjewel}
\end{figure}

\end{enumerate}

The aforementioned changes to make MCTS multiplayer have been described in previous literature as MCTS-max\textsuperscript{n} \cite{multimcts}.

\section{Experiments \& Results}
We define several metrics of success in training effective agents:
\begin{itemize}
    \item \textbf{Does the neural network successfully converge?} A stable, decreasing loss function indicates training is proceeding as anticipated. Divergence likely indicates the network is low capacity or over-regularized, as it cannot explain the growing variance of experience data.
    \item \textbf{Does the agent outperform a MCTS given the same number of rollouts?} Since the AlphaZero agent starts off as a standard MCTS agent and improves from there, it should outperform a MCTS agent given the same number of rollouts per turn. This experiment tests if the experimental agent (AlphaZero) outperforms the control agent (MCTS).
    \item \textbf{Does the agent outperform a MCTS given more rollouts (up to a point)?} Since AlphaZero's heuristic enables it to efficiently search the game tree, it should perform as well as or better than some MCTS agents that are given additional rollouts. Since AlphaZero only plays against itself during training, this experiment tests the generality of the learned strategy.
    \item \textbf{Does the agent outperform a human?} There are no human experts for the games we have created, but victory against a competent human opponent confirms that a reasonably strong and general strategy was learned.
\end{itemize}

Using these criteria to evaluate multiplayer agents, we train AlphaZero to play multiplayer versions of Tic-Tac-Toe and Connect 4.

\paragraph{Testbed.}
We have implemented multiplayer AlphaZero entirely in Python 3 using PyTorch. Unlike DeepMind's AlphaZero, we do not parallelize computation or optimize the efficiency of our code beyond vectorizing with numpy. All experiments were run on a desktop machine containing an i9-9900k processor and an RTX 2080 Ti GPU. Our biggest limitation was compute - DeepMind trained AlphaZero to master Go in 13 days with 5000 first generation TPUs and 16 second-generation TPUs, but with our hardware, that result would take years to replicate \cite{deepmind}. For this reason, we experiment on multiplayer games with small state and action spaces to make this project feasible. Even on these simpler games, training takes over 15 hours. Future research with access to more compute can expand on our results by evaluating performance on more complex multiplayer games.

\paragraph{Hyperparameters.}
The same hyperparameters are used across all games and experiments. For the neural network, we use a squeeze-and-excitation model, which has been shown to outperform existing DCNN architectures by modeling channel interdependencies with only a slight increase to model complexity \cite{squeeze}. The specific SENet architecture used in this project consists of 8 SE-PRE blocks and two heads (value and policy).

\begin{table}[h]
  \caption{Hyperparameters}
  \label{hyperparams}
  \centering
  \begin{tabular}{lll}
    \toprule
    \cmidrule(r){1-2}
    Hyperparameter     & Ours     & DeepMind \\
    \midrule
    Network & SENet & ResNet  \\
    L2 regularization & 1e-4 & 1e-4\\
    Batch size & 64 & 2048\\
    Optimizer & ADAM & SGD + Momentum\\
    Learning rate & 1e-3 & 1e-2 $\longrightarrow$ 1e-4 \\
    Replay buffer size & $\infty$ & ?\\
    $c_{puct}$ & 3.0 & ? \\
    Dirichlet $\alpha$ & 1.0 &  Depends on game \\
    MCTS rollouts per turn (``computation'') & 50 & 800\\
    \bottomrule
  \end{tabular}
\end{table}

\subsection{Multiplayer Tic-Tac-Toe}
\begin{figure}[h]
  \begin{subfigure}[t]{.45\textwidth}
  \centering
    \includegraphics[width=.9\textwidth]{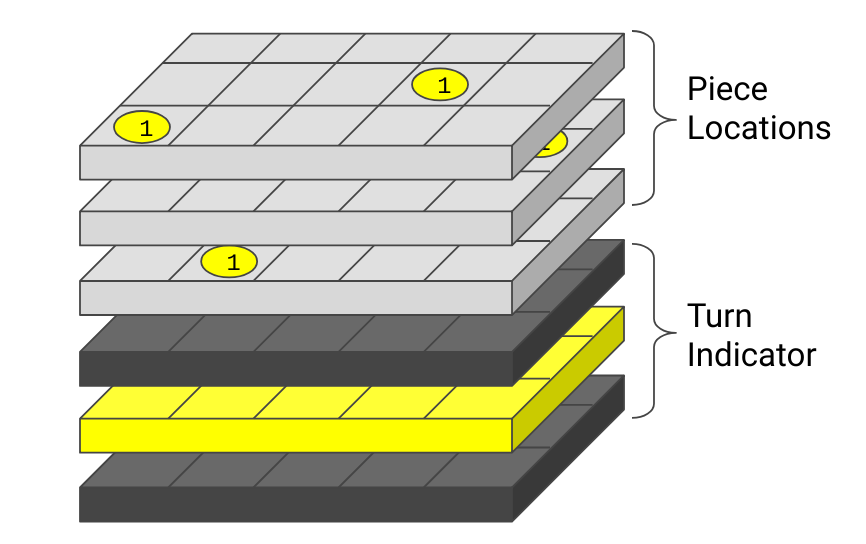}
    \caption{The state representation of a Tic-Tac-Mo board passed into the neural network. Size is 3x5x6. Each player owns one "piece location" plane and "turn indicator" plane.}
    \label{fig:ttm_state}
  \end{subfigure}\hfill
  \begin{subfigure}[t]{.45\textwidth}
  \centering
    \includegraphics[width=.9\textwidth]{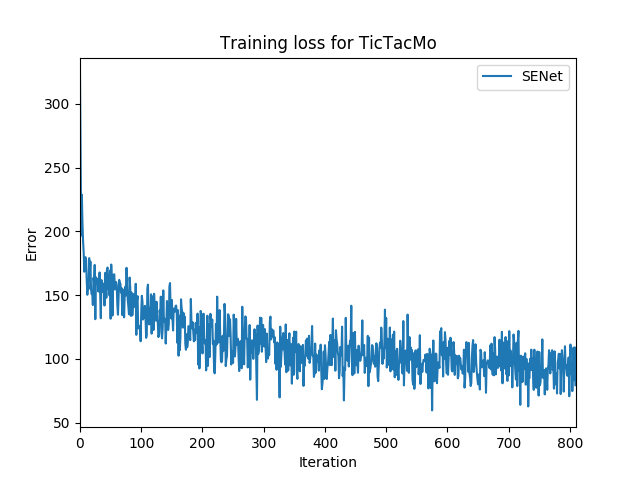}
    \caption{The loss of our SENet steadily converges.}
    \label{fig:ttm_loss}
  \end{subfigure}
  \caption{State representation and loss curve for Tic-Tac-Mo.}
\end{figure}

Our multiplayer Tic-Tac-Toe game, dubbed ``Tic-Tac-Mo,'' adds an additional player to Tic-Tac-Toe but keeps the 3-in-a-row win condition. To make games more complicated, the size of the board is expanded to be 3x5 instead of 3x3. Games can therefore last up to 15 turns. Players receive a score of 1 for a win, 0 for a tie, and -1 for a loss. The state representation of the board fed into the neural network is depicted in figure \ref{fig:ttm_state}. We trained the AlphaZero algorithm by having it play Tic-Tac-Mo against itself for about 18 hours.

\paragraph{Does the neural network successfully converge?}
The loss curve for the underlying heuristic network is shown in figure \ref{fig:ttm_loss}. The loss decreases and stabilizes as AlphaZero goes through more iterations.

\begin{figure}[h]
  \begin{subfigure}[t]{.45\textwidth}
  \centering
    \includegraphics[width=.9\textwidth]{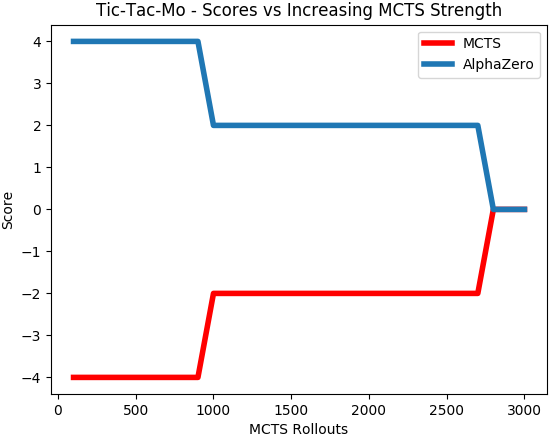}
    \caption{AlphaZero and opponent scores accumulated over six games as opponent rollouts increase. AlphaZero's rollouts remain fixed at 50, while its MCTS opponents use an increasing number of rollouts. The two MCTS agents have identical performance across each match.}
    \label{fig:ttm_strength}
  \end{subfigure}\hfill
  \begin{subfigure}[t]{.45\textwidth}
  \centering
    \includegraphics[width=.9\textwidth]{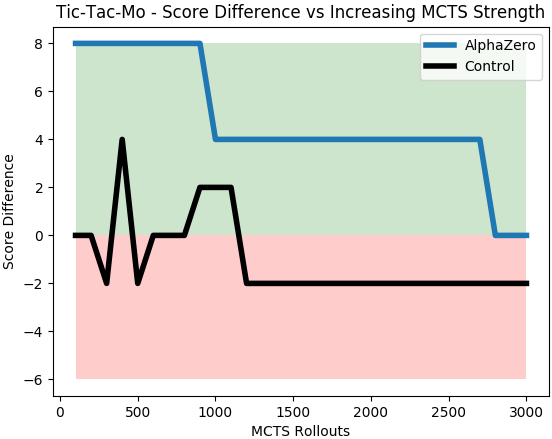}
    \caption{Score difference as opponent rollouts increase for AlphaZero and a control MCTS using the same number of rollouts. Score difference is the difference between our score and the opponent's score. Score differences less than 0 indicate more games lost than won, score differences greater than 0 indicate more games won than lost, and score differences of 0 indicate equal wins and losses.}
    \label{fig:ttm_control}
  \end{subfigure}
  \caption{Tic-Tac-Mo experiments againsts MCTS opponents of increasing strength.}
\end{figure}

\paragraph{Does the agent outperform a MCTS given more rollouts (up to a point)?} We compare the scores between AlphaZero and MCTS opponents of increasing strength. Here, ``increasing strength'' means that after each match, MCTS gets more rollouts to search the game tree, while AlphaZero's computation remains fixed at 50 rollouts. Each game pits 2 MCTS agents of equal strength against our AlphaZero agent. For each match, a total of 6 games is played between the same opponents - one game for each permutation of players to break any advantages from going first, second, or third. Figure \ref{fig:ttm_strength} plots the scores of AlphaZero against the scores of MCTS for each match. We find AlphaZero convincingly defeats MCTS agents that have few rollouts, but score starts to converge as MCTS strength increases.

\paragraph{Does the agent outperform a MCTS given the same number of rollouts?} A control MCTS agent \textit{given the same number of rollouts as AlphaZero} (50 rollouts) is \textit{also} played against MCTS opponents of increasing strength. The difference in points between the control and its opponents is plotted alongside the difference in points between AlphaZero and its opponents (figure \ref{fig:ttm_control}). We find AlphaZero always performs better than the control MCTS when playing against the same opponents.

\paragraph{Does the agent outperform a human?} We had several graduate students play Tic-Tac-Mo against two AlphaZero agents (table \ref{ttm-humans}). We described the game to this small group of students and gave them a chance to practice before having each student play 6 games against AlphaZero opponents (order of players was permuted each round). For these AlphaZero agents, we increased their rollouts to 500 - if the learned heuristic function is general enough, it should scale given more search time and develop extremely strong strategies. In total, AlphaZero tied 42\% of the games and won 58\% of the games as shown in table \ref{ttm-humans}. \\

\begin{table}[h]
  \caption{Summary of human performance against AlphaZero, Tic-Tac-Mo.}
  \label{ttm-humans}
  \centering
  \begin{tabular}{llll}
    \toprule
    \cmidrule(r){1-3}
    Human & Wins & Ties & Losses \\
    \midrule
     Human 1   & 0 & 1 & 5\\
     Human 2   & 0 & 4 & 2\\
     Human 3   & 0 & 3 & 3\\
     Human 4   & 0 & 2 & 4\\
     \midrule
     Totals & 0 & 10 & 14 \\
    \bottomrule
  \end{tabular}
\end{table}

\paragraph{Results.}
Multiplayer AlphaZero for Tic-Tac-Mo suffered no defeats against computer or human opponents. The decreasing network loss function is an indication that the network trained successfully, continually improving its estimates of policy and value while incorporating new experience. With just 50 rollouts, AlphaZero has equivalent performance to at least a 3000-rollout MCTS - and can pick a move in a fraction of the time. Our control experiment indicates that the learned heuristic is necessary and useful, leading us to believe our multiplayer AlphaZero algorithm successfully encoded knowledge of the game into its heuristic, creating a powerful Tic-Tac-Mo agent.

\subsection{Multiplayer Connect 4}
\begin{figure}[h]
  \begin{subfigure}[t]{.45\textwidth}
  \centering
    \includegraphics[width=.9\textwidth]{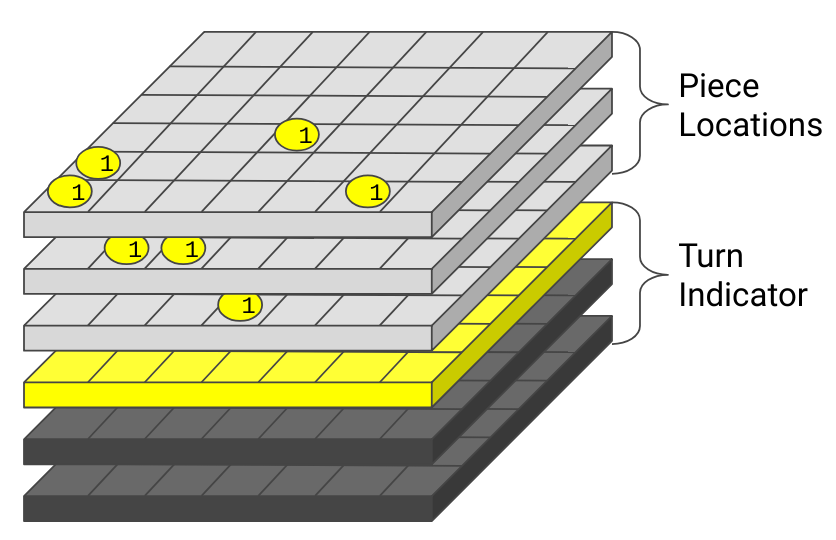}
    \caption{The state representation of a Connect 3x3 board passed into the neural network. Size is 6x7x6. Each player owns one "piece location" plane and "turn indicator" plane.}
    \label{fig:c3_state}
  \end{subfigure}\hfill
  \begin{subfigure}[t]{.45\textwidth}
  \centering
    \includegraphics[width=.9\textwidth]{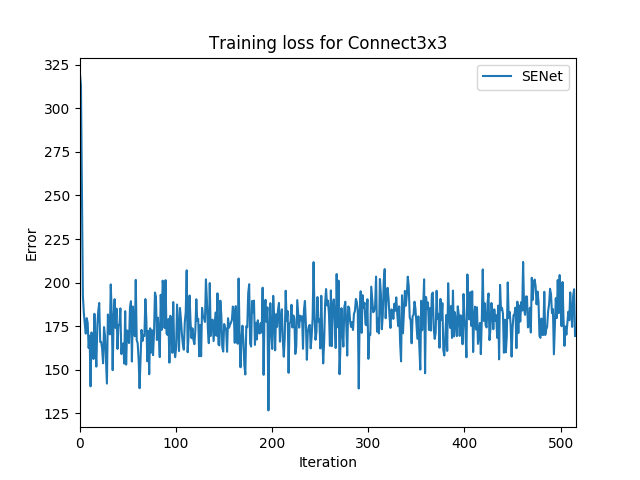}
    \caption{The loss of our SENet is stable.}
    \label{fig:c3_loss}
  \end{subfigure}
  \caption{State representation and loss curve for Connect 3x3.}
\end{figure}

Our multiplayer Connect 4 game dubbed ``Connect 3x3'' adds an additional player to the game and changes the win condition to 3-in-a-row instead of 4-in-a-row. The size of the board remains 6x7. We believe Connect 3x3 to be a harder game to learn than Tic-Tac-Mo, as games can last up to 42 turns as opposed to 15, so the game tree is much deeper. Players receive a score of 1 for a win, 0 for a tie, and -1 for a loss. The state representation of the board fed into the neural network is depicted in figure \ref{fig:c3_state}. We trained the algorithm by having it play Connect 3x3 against itself for about 18 hours.

\paragraph{Does the neural network successfully converge?}
The loss curve for the network is shown in figure \ref{fig:c3_loss}. Error does not steadily decrease over time but remains relatively stable.

\begin{figure}[h]
  \begin{subfigure}[t]{.45\textwidth}
  \centering
    \includegraphics[width=.9\textwidth]{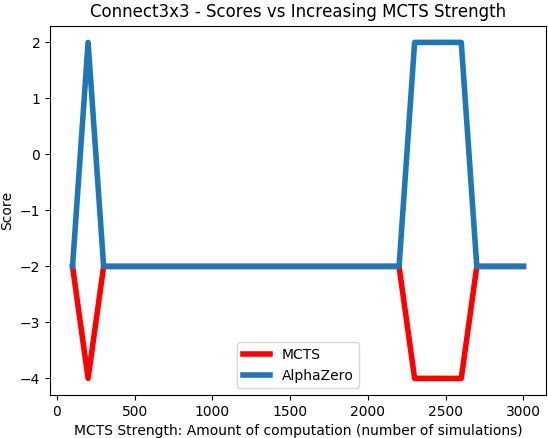}
    \caption{AlphaZero and opponent scores over six games as opponent rollouts increase.}
    \label{fig:c3_strength}
  \end{subfigure}\hfill
  \begin{subfigure}[t]{.45\textwidth}
  \centering
    \includegraphics[width=.9\textwidth]{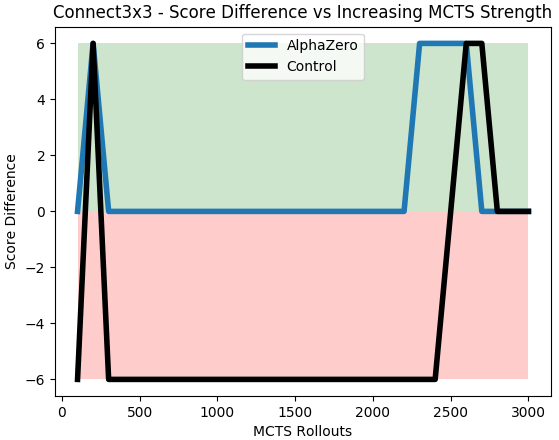}
    \caption{Score difference as opponent rollouts increase for AlphaZero and a control.}
    \label{fig:c3_control}
  \end{subfigure}
  \caption{Connect 3x3 experiments againsts MCTS opponents of increasing strength.}
\end{figure}

\paragraph{Does the agent outperform a MCTS given more rollouts (up to a point)?}
We run the same experiment as described in Tic-Tac-Mo, but now for Connect 3x3 (figure \ref{fig:c3_strength}). Like with Tic-Tac-Mo, AlphaZero ties or outperforms each MCTS opponent. Unlike Tic-Tac-Mo, we do not see a converging score gap between MCTS and AlphaZero - instead score appears to oscillate as MCTS increases in strength. 

\paragraph{Does the agent outperform a MCTS given the same number of rollouts?} We again run a control MCTS agent and compare it to our AlphaZero agent (figure \ref{fig:c3_control}). From these results, we see that the control is mostly losing to stronger MCTS agents while AlphaZero maintains a non-negative score difference. In general, AlphaZero outperforms the control, however there is one blip where the control outperforms AlphaZero.

\paragraph{Does the agent outperform a human?} Finally, we had the same group of graduate students now play against two Connect 3x3 AlphaZero agents (table \ref{c3-humans}). In total, AlphaZero won 79\% of the games and lost 21\% of the games.  \\

\begin{table}[h!]
  \caption{Summary of human performance against AlphaZero, Connect 3x3.}
  \label{c3-humans}
  \centering
  \begin{tabular}{llll}
    \toprule
    \cmidrule(r){1-3}
    Human & Wins & Ties & Losses \\
    \midrule
     Human 1   & 0 & 0 & 6\\
     Human 2   & 0 & 0 & 6\\
     Human 3   & 3 & 0 & 3\\
     Human 4   & 2 & 0 & 4\\
     \midrule
     Totals & 5 & 0 & 19 \\
    \bottomrule
  \end{tabular}
\end{table}

\paragraph{Results.} 

Multiplayer AlphaZero trains a strong agent to play Connect 3x3 which wins or ties most games. However, unlike with Tic-Tac-Mo, we do have humans who are able to defeat AlphaZero, and a case where the control MCTS agents outperforms AlphaZero. Both of these measurements indicate that AlphaZero did \textit{not} perfect its neural network board-quality heuristic and master Connect 3x3.

But the learned heuristic, though fallible, still successfully encodes knowledge of the game into search. With just 50 rollouts, AlphaZero meets or beats its MCTS opponents, and typically outperforms a control MCTS agent given the same number of rollouts. And though the loss function is not decreasing, it does not diverge either. Since training data is continually added to the replay buffer, this indicates knowledge is being incorporated and generalized into the network.

Our results from Connect 3x3 indicate that the overall multiplayer AlphaZero strategy works, but more hyperparameter tuning is needed to truly master complex games.

\section{Conclusion}

In this paper we propose a novel modification to the AlphaZero algorithm that enables it to train multiplayer agents through self-play. Our experiments show that AlphaZero can be successfully applied to multiplayer games, but more careful hyperparameter tuning is necessary to achieve stronger agents. We define measures of success that can be applied in future AlphaZero research, and create an independent AlphaZero reimplementation with multiplayer modification. Given more computation, future work should include experiments on games with more players, more board states, and more actions. Other research directions might investigate the effectiveness of AlphaZero when other constraints such as \textit{zero-sum game}, \textit{deterministic game}, or \textit{perfect information game} are lifted.

\bibliographystyle{plainnat}
\bibliography{references}

\end{document}